\documentclass[letterpaper,11pt]{report}

\pdfoutput=1

\usepackage{graphicx}
\usepackage{latexsym}
\usepackage{makeidx}
\usepackage{url}
\usepackage[latin1]{inputenc}
\usepackage{listings}

\makeindex

\newcommand{\xf}[1]{Figure~\ref{#1}}

\newcommand{\xt}[1]{Table~\ref{#1}}

\newcommand{\rmi}{{RMI\index{RMI}}}

\newcommand{\corba}{{CORBA\index{CORBA}}}

\newcommand{\java}{{Java\index{Java}}}

\newcommand{\file}[1]{\url{#1}\index{Files!#1}}
\newcommand{\tool}[1]{\texttt{#1}\index{Tools!#1}}

\newcommand{\api}[1]{\texttt{#1}\index{API!#1}}
\newcommand{\apipackage}[1]{\url{#1}\index{API!Packages!#1}\index{Packages!#1}}

\newcommand{\marf}[0]{MARF\index{Tools!MARF}\index{Libraries!MARF}}

\newcommand{\lucidL}[1]{{$\mathit{Lucid}$}($L$) }

		{}

\def\myvert{\raise 2.27pt \hbox{\vrule depth 0pt height 8pt width 0.2mm}}
\def\myarrow{\hspace*{0.43mm}%
             \raise 2.29pt\hbox{\vrule depth 0pt height 8pt width 0.16mm}%
             \hspace*{-0.32mm}%
             $\longrightarrow$
             \ %
             }

\begin{document}
\pagestyle{headings}

\title
{
	{\bf On Design and Implementation of Distributed Modular Audio Recognition Framework}\\
	Requirements and Specification Design Document\vfill
	{\small Concordia University}\\
	{\small Department of Computer Science and Software Engineering}\vfill
}
\author
{
	MARF Research \& Development Group\thanks{http://marf.sf.net}:\\
	Serguei A. Mokhov\\
	\texttt{mokhov@cse.concordia.ca}
}
\date{\vfill Montreal, Quebec, Canada\\August 12, 2006}
\maketitle

\pagenumbering{roman}
\tableofcontents
\clearpage
\pagenumbering{arabic}

\listoffigures
\listoftables

\chapter{Executive Summary}

This chapter highlights some details for the inpatient readers, while
the rest of the document provides a lot more details.

\section{Brief Introduction and Goals}

\begin{itemize}
\item
An open-source project -- {\marf} (\url{http://marf.sf.net}, \cite{marf}), which stands
for Modular Audio Recognition Framework -- originally designed for the
pattern recognition course.

\item
{\marf} has several applications. Most revolve around its recognition
pipeline -- sample loading, preprocessing, feature extraction, training/classifcation.
One of the applications, for example, is Text-Independed Speaker
Identification Application. The pipeline and the application, as they stand, are purely
sequential with even little or no concurrency when processing a bulk of voice
samples.

\item
The classical {\marf}'s pipeline is in \xf{fig:pipeline-flow}.
The goal of this work is to distribute the shown stages of the pipeline as
services as well as stages that are not directly present in
the figure -- sample loading, front-end application service (e.g.
speaker identification service, etc.) and implement some disaster
recovery and replication techniques in the distributed system.

\item
In \xf{fig:pipeline-net} the design of the distributed version of the pipeline is
presented. It indicates different levels of basic front-ends, from
higher to lower which client applications may invoke as well
as services may invoke other services through their front-ends
while executing in the pipeline mode. The back-ends are in charge
of providing the actual servant implementations as well as the
features like primary-backup replication, monitoring, and disaster
recovery modules through delegates.

\end{itemize}

\section{Implemented Features So Far}

\begin{itemize}
\item
As of this writing the following are implemented.
Most, but not all modules work:

\item
Out of the following six services:

\begin{enumerate}
\item SpeakerIdent Front-end Service (invokes MARF)
\item MARF Pipeline Service (invokes the remaining four)
\item Sample Loader Service
\item Preprocessing Service
\item Feature Extraction Service (may invoke Preprocessing for preprocessed sample)
\item Classification (may invoke Feature Extraction for features)
\end{enumerate}

all the six work in the stand-alone and pipelined modes in {\corba}, {\rmi},
and WS.

At the demo time, the {\rmi} and as
a consequence in Web Services implementation of the Sample Loader and
Preprocessing stages were not functional (other nodes were, but
could not work as a pipeline) because of the design
flaw in the {\marf} itself (the \api{Sample} class data structure
while itself was \api{Serializable}, one of its members, that
inherits from a standard Java class, has non-serializable members
in the parent) causing marshalling/unmarshalling to fail. This has
been addressed until after demo.

\item
There are three clients: one for each communication technology
type (CORBA, WS, RMI).

\item
MARF vs. CORBA vs. RMI object adapters to convert serializable objects
understood by the technologies to the MARF native and back.
\end{itemize}

\section{Some Design Considerations}

\begin{itemize}
\item
For WS there are no remote object references, so
a class was created called \api{RemoteObjectReference}
encapsulating nothing but a type (\api{int}) and
an URL (\api{String}) as a reference that can be
passed around modules, which can later use it
to connect (using \api{WSUtils}).

\item
All communication modules rely on their delegates
for business and mosf of the transaction logic,
thus remapping remote operations to communication-technology
idependent logic and enabling cross-technology
commuincation through message passing.  There are two types of
delegates -- basic and recoverable. The basic delegates just
merely redirect the business logic and provide basis for
transaction logs while not actually implementing the transaction
routines. They don't endure the transactions overhead and just allow
to test the distributed business logic. The recoverable delegates
are extension of the basic with the transactionaly on top of the
basic operations.

\item
All modules also have utility classes like \api{ORBUtils}, \api{RMIUtils},
and \api{WSUtils}. These are used by the distributed modules for common
registration of services and their look up. Due to
the common design, these can be looked up at run-time
through a reflection by loading the requested module classes. The utility
modules are also responsible for loading the initial service location
information from the \file{dmarf-hosts.properties} when available.
\end{itemize}

\section{Transactions, Recoverablity, and WAL Design}

\begin{itemize}
\item
Write-Ahead Log (WAL) consists of entries called ``Transactions''.
The idea is that you write to the log first, ahead of committing
anything, and once write call (dump) returns, we commit the transaction.

\item
A \api{Transaction} is a data structure maintaining transaction ID (\api{long}),
a filename of the object (not of the log, but where the object is normally
permanently stored to distinguish different configurations), the \api{Serializable} value itself (a \api{Message},
\api{TrainingSet}, or an entire \api{Serializable} business-logic module), and timestamps.

\item
The WAL's max size is set to empirical 1000 entries before clean up is needed.
Advantage of keeping such entries is to allow a future feature
called point-in-time recovery (PITR), backup, or replication.

\item
MARF-specific note: since MARF core operations are treated
as kind of a business logic black box, the ``transactions'' are similar to the
``before'' and ``after'' snapshots of serialized data (maybe a design
flaw in MARF itself, to be determined).

\item
Checkpointing in the log is done periodically, by default every second.
A checkpoint is set to be a transaction ID latest committed.
Thus, in the event of a crash, to recover, only committed transactions
with the ID greater than the checkpoint are recovered.
\end{itemize}

\section{Configuration and Deployment}

All {\corba}, {\rmi}, and WS use a \file{dmarf-hosts.properties}
files at startup if available to locate where the
other services are and where to register themselves.

Web Services have Tomcat context XML files for hosting
as well as \file{web.xml} and related WSDL XML files.

All such things are scripted in the GNU Make \cite{gmake} \file{Makefile}
and Ant \cite{ant} \file{dmarf-build.xml} makefiles.

\section{Testing}

A Makefile target  \texttt{marf-client-test} for a single wave file and a
\file{batch.sh} shell script
test mostly CORBA pipeline with 295 testing samples and
31 testing wave samples x 4 training configs x 16 testing
configs.

The largest demo experiment invovled only four machines
in two different buildings running the 6 services and a client
(a some machines ran more than one service of each kind).
Killing any of the single services in batch mode and then
restarting it, recovered the ability of a pipeline to
operate normally.

\section{Known Issues and Limitations}

\begin{itemize}
\item
After long runs of all six CORBA services on the same machine
runs out of file (and socket) descriptors reaching default
kernel limits. (Probably due to large number of log files
opened and not closed while the containing JVM does not exit
and which accumulate over time after lots of rigorous testing).

\item
Main MARF's design flaws making the pipeline rigid and less
concurrent (five-layer nested transaction, see \api{startRecognitionPipeline()}
of \api{MARFServerCORBA}, \api{MARFServerRMI}, or \api{MARFServerWS} for examples.

\item
Transaction ID ``wrap-around'' for long-running system and
transactions with lots of message passing and other operations.
MARF does a lot of writes (dumps) and long-running servers
have a potential to have their transaction IDs be recycled
after an overflow. At the time of this writing, there is no
an estimate of how log it might take when this happens.

\item
All services are single-threaded in the proof-of-concept
implementation, so the concurrency is far from being fully
exploited per server instance. This is to be overcome in
the near future.
\end{itemize}

\section{Partially Implemented Planned Features}

\begin{itemize}
\item
WAL logging and recovery.

\item
Message passing (for gossip, TPC or UDP + FIFO) is to be added
to the basic delegates.

\item
Application and Status Monitor GUI -- the rudiments are there, but
not fully integrated yet.
\end{itemize}

\section{NOT Implemented Planned Features}

\begin{itemize}
\item
Primary-backup replication with a ``warm stanby''.
\item
Lazy, gossip-based replication for Classification training sets.
\item
Two-phase commit for nested MARF Service transactions (covering
the entire pipeline run.
\item
Distributed System-ware NLP-related applications.
\item
Thin test clients and their GUI.
\end{itemize}

\section{Conclusion}

This proof-of-concept implementation of Distributed MARF
has proven a possibility for the pipeline stages and not only
to be executed in a pipeline and stand-alone modes on several
computers. This can be useful in providing any of the mentioned services
to clients that have low computational power or no required environment
to run the whole pipeline locally or cannot afford long-running processing
(e.g. collecting samples with a laptop or any mobile device and submitting
them to the server). Additionally, there were discovered some show-stopping design flaws in the
classical MARF itself that have to be corrected, primarily related to
the storage and parameter passing among modules.

\section{Future Work}

Address the design flaws, limitations, and not-implemented
features and release the code (for future improvements).
{\em You} may volunteer to help to contribute these ;-) as well as addressing
the bugs and limitations when there is a time and desire.
Please email to \url{mokhov@cse.concordia.ca} if you are intrested
in contributing to the Distributed MARF project.


\clearpage


\chapter{Introduction}
\index{Introduction}

$Revision: 1.3 $

This chapter briefly presents the purpose and the scope of the work on the Distributed {\marf} project
with a subset of relevant requirements, definitions, and acronyms.
All these aspects
are detailed to some extent later through the document.
The application ideas in small
part are coming from \cite{disysconcepts05, java-rmi, java-corba-idl, java-networking, java-webservices, marf, mokhov-dsb}.

\section{Requirements}
\label{sect:requirements}
\index{Introduction!Requirements}

I have an open-source project -- {\marf} (\url{http://marf.sf.net}, \cite{marf}), which stands
for Modular Audio Recognition Framework. Originally designed for the
pattern recognition course back in 2002, it had addons from other courses
I've taken and maintained and released it relatively regularly.

{\marf} has several applications. Most revolve around its recognition
pipeline -- sample loading, preprocessing, feature extraction, training/classifcation.
One of the applications, for example is Text-Independed Speaker
Identification. The pipeline and the application as they stand are purely
sequential with even little or no concurrency when processing a bulk of voice
samples. Thus, the purpose of this work is to make the pipeline distributed and run on a cluster
or a just a set of distinct computers to compare with the traditional version and add disaster recovery and service replication,
communication technology indepedence, and so on.

\begin{figure}
	\centering
	\includegraphics[width=\textwidth]{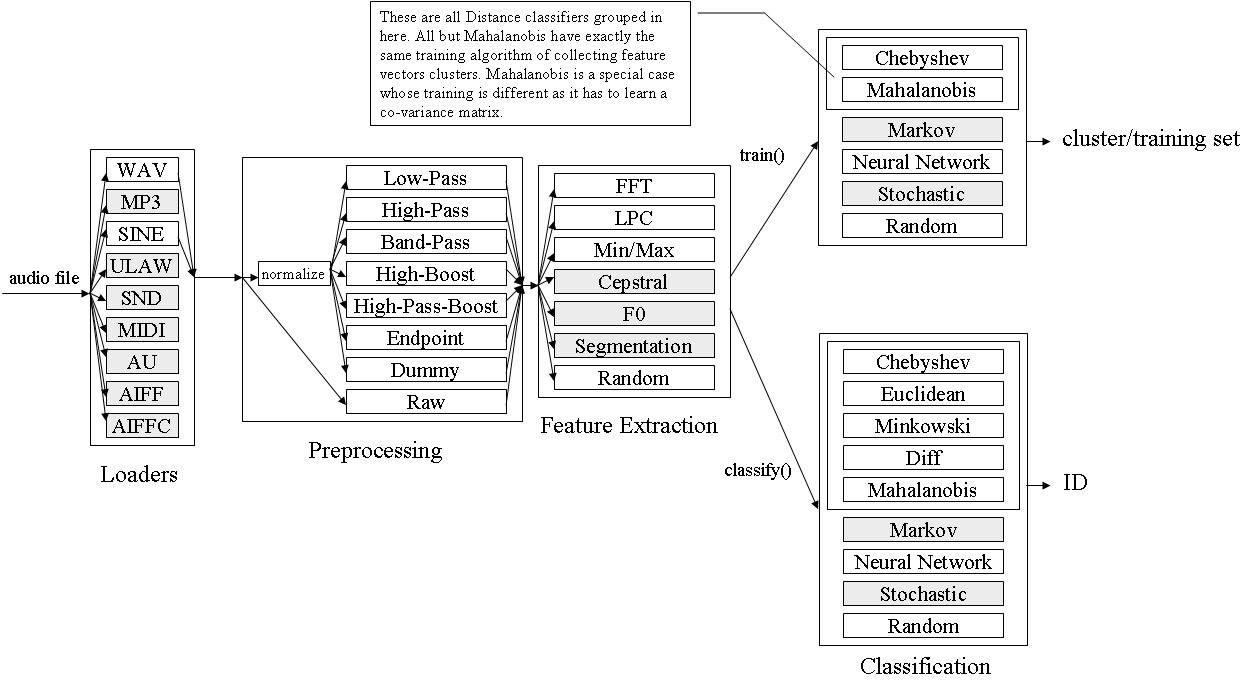}
	\caption{The Core MARF Pipeline\index{MARF!Core Pipeline} Data Flow}
	\label{fig:pipeline-flow}
\end{figure}

The classical {\marf}'s pipeline is in \xf{fig:pipeline-flow}.
The goal of this work is to distribute the shown stages of the pipeline as
services as well as stages that are not directly present in
the figure -- sample loading, front-end application service (e.g.
speaker identification service, etc.) and implement some disaster
recovery and replication techniques in the distributed system.

\begin{figure}
	\centering
	\includegraphics[width=\textwidth]{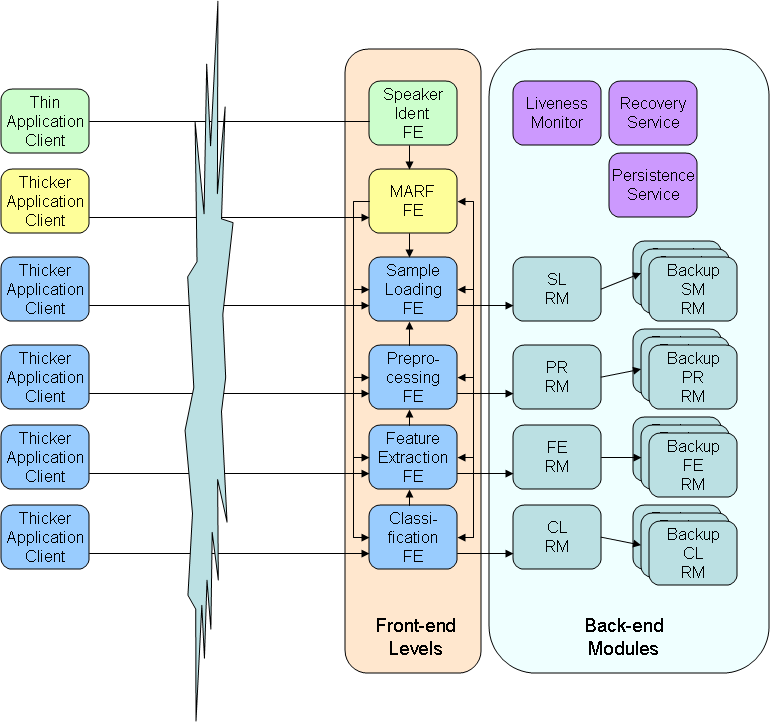}
	\caption{The Distributed MARF Pipeline\index{MARF!Distributed Pipeline}}
	\label{fig:pipeline-net}
\end{figure}

In \xf{fig:pipeline-net} the distributed version of the pipeline is
presented. It indicates different levels of basic front-ends, from
higher to lower, which a client application may invoke as well
as services may invoke other services through their front-ends
while executing in a pipeline-mode. The back-ends are in charge
of providing the actual servant implementations as well as the
features like primary-backup replication, monitoring, and disaster
recovery modules.

There are several distributed services, some are more general, and
some are more specific. The services can and have to intercommunicate.
These include:

\begin{itemize}
\item
General MARF Service that exposes {\marf}'s pipeline to clients and other
services and communicates with the below.

\item
Sample Loading Service knows how to load certain file or stream types (e.g. WAVE)
and convert them accordingly for further preprocessing.

\item
Preprocessing Service accepts incoming voice or text samples and does
the requested preprocessing (all sorts of filters, normalization, etc.).

\item
Feature Exraction Service accepts data, presumably preprocessed, and
attempts to extract features out of it given requested algorithm (out of
currently implemented, like FFT, LPC, MinMax, etc.)
and may optionally query the preprocessed data from the Preprocessing
Service.

\item
Classifcation and Training Service accepts feature vectors
and either updates its database of training sets or performs
classification against existing training sets. May optionall query
the Feature Extraction Service for the features.

\item
Natural Language Processing Service accepts natural language texts and
performs also some statistical NLP operations, such as probabilistic
parsing, Zipf's Law stats, etc.
\end{itemize}

Some more application-specific front-end services (that are based on the
existing currently non-distributed apps) include but not limited to:

\begin{itemize}
\item
Speaker Identification Service (a front-end) that will communicate with the MARF
service to carry out application tasks.

\item
Language Identification Service would communicate with MARF/NLP for the similar purpose.

\item
Some others (front-ends for Zipf's Law, Probabilistic Parsing, and test
applications).
\end{itemize}

The clients are so-called ``thin'' clients with GUI or a Web Form
allowing users to upload the samples for training/classification and set
the desired configuration for each run, either for individual samples or
batch.

Like it was done in the
Distributed Stock Broker \cite{mokhov-dsb}, the architecture is general and usable enough to enable one
or more services using CORBA, RMI, Web Services (WS), Jini, JMS, sockets,
whatever (well, actually, Jini, JMS were not implemented in either applications, but it is
not a problem to add with little or no ``disturbance'' of the rest of the
architecture).

\section{Scope}
\index{Introduction!Scope}

In the Distributed {\marf}, if any pipeline stage process
crashes access to information about the pending transactions and computatiion in module is not only lost
while the process remains unavailable but can also be lost forever.

Use of a message
logging protocol is one way that a module could recover information concerning that
module's data after a faulty processor has been repaired.
A WAL message-logging protocol is developed for DMARF.
The former is for the disaster recovery of uncommitted transactions and to
avoid data loss. It also allows for backup replication and point-in-time recovery
if WAL logs are shipped off to a backup storage or a replica manager and can be
used to reconstruct the replica state via gossip or any other replication scheme.

The DMARF is also extended by adding a ``warm standby''. The ``warm standby'' is a MARF module that
is running in the background (normally on a different machine), receiving operations from
the primary server to update its state and hence ready to jump in if the primary server
fails. Thus, when the primary server receives a request from a client which will change its
state, it sends the request to the backup server, performs the request, receives the response
from the backup server and then sends the reply back to the client.
The main purpose of the ``warm stand by'' is
to minimise the downtime for subsequent transactions while the primary is in disaster recovery.
The primary and
backup servers communicate using either the reliable
TCP protocol (over WAN) or a FIFO-ordered UDP on a LAN.
Since this is a secondary feature and the load in this project will be more
than average, we simply might not have time to do and debug this stuff to
be reliable over UDP, so we choose TCP do it for us, like we
did in StockBroker Assignment 2. IFF we have time, we can try to make a FIFO
UDP communication.

\begin{itemize}
\item
Design and implement the set of required interfaces in RMI, CORBA, and WS
for the main MARF's pipeline stages to run distributedly, including any possible
application front-end and client applications.

\item
Assuming that processor failures are benign (i.e. crash failures) and not
Byzantine, analysis of the classical MARF was done to determine the information necessary for the proper
recovery of a MARF module (that is, content of the log) and the design of the ``warm standby'' replication system.

\item
Modify MARF implementation so that it logs the required
information using the WAL message-logging protocol.

\item
Design and implement a recovery module which restarts a MARF module
using the log so that the restarted module can process subsequent requests for the
various operations.

\item
Design and implement the primary server which receives requests from
clients, sends the request to the backup server, performs the request, and sends the
response back to the client only after the request has been completed correctly both in
the primary and the backup servers. When the primary notices that the backup does not
respond within a reasonable time, it assumes and informs the MARF monitor that the
backup has failed so that a new backup server can be created and initialized.

\item
Design and implement a monitor module which periodically checks
the module process and restarts it if necessary.
application.  This monitor initializes the primary and backup
servers at the beginning, creates and initializes a backup server when the primary fails
(and the original backup server takes over as the primary), and creates and initializes a
backup server when the original backup server fails.

\item
Design and implement the backup server which receives requests from the
primary, performs the request and sends the reply back to the primary. If the backup
server does not receive any request from the primary for a reasonable time, it sends a
request to the primary to check if the latter is working. If the primary server does not
reply in a reasonable time, the backup server assumes that the primary has failed and
takes over by configuring itself as the primary so that it can receive and handle all
client requests from that point onwards; and also informs the broker monitor of the
switch over so that the latter can create and initialize another backup server.

\item
Integrate all the modules properly, deploy the application on a local area
network, and test the correct operation of the application using properly designed test
runs. One may simulate a process failure by killing that process from the command line
while the application is running.

\end{itemize}

\section{Definitions and Acronyms}
\index{Introduction!Definitions and Acronyms}

\begin{description}
\index{Introduction!Definitions and Acronyms}

\item[API] Application Programmers Interface -- a common convenience collection of objects,
	methods, and other object members, typically in a library, available for an application
	programmer to use.

\item[CORBA] Common Object Request Broker Architecture -- a language model independent platform
	for distributed execution of applications possibly written in different languages, and, is,
	therefore, heterogeneous type of RPC (unlike Java RMI, which is Java-specific).

\item[HTML] HyperText Markup Language -- a tag-based language for defining the layout of web pages.

\item[IDL] Interface Definition Language -- a CORBA interface language to ``glue'' most common types
	and data structures in a specific programming language-independent way. Interfaces written in IDL
	are compiled to a language specific definitions using defined mapping between constructs in IDL
	and the target language, e.g. IDL-to-Java compiler (\tool{idlj}) is used for this purpose in this
	assignment.

\item[CVS] Concurrent Versions System -- a version and revision control system to manage
	source code repository.

\item[DSB] Distributed Stock Broker application.

\item[DMARF] Distributed {\marf}.

\item[J2SE] Java 2 Standard Edition.

\item[J2EE] Java 2 Entreprise Edition.

\item[JAX-RPC] Java XML-based RPC way of implementing Web Services.

\item[JAX-WS] The new and re-engineered way of Java Web Services implementation as opposed
	to the older and being phased-out Java XML-RPC.

\item[JDK] The Java Development Kit. Provides the JRE and a set of tools (e.g. the \tool{javac}, \tool{idlj}, \tool{rmic}
	compilers, \tool{javadoc}, etc.) to develop	and execute Java applications.

\item[JRE] The Java Runtime Environment. Provides the JVM and required libraries to execute
	Java applications.

\item[JVM] The Java Virtual Machine. Program and framework allowing the execution of program
	developed using the Java programming language.

\item[MARF] Modular Audio Recognition Framework \cite{marf} has a variety of useful general
	purpose utility and storage modules employed in this work, from the same author.

\item[RMI] Remote Method Invocation -- an object-oriented way of calling methods of objects
	possibly stored remotely with respect to the calling program.

\item[RPC] A concept of Remote Procedure Call, introduced early by Sun, to indicate that
	an implementation certain procedure called by a client may in fact be located remotely
	from a client on another machine

\item[SOAP] Simple Object Access Protocol -- a protocol for XML message exchange over HTTP
	often used for Web Services.

\item[STDOUT] Standard output -- an output data stream typically associated with a screen.

\item[STDERR] Standard error -- an output data stream typically associated with a
	screen to output error information as opposed to the rest of the output sent to
	STDOUT.

\item[WS] Web Services -- another way of doing RPC among even more heterogeneous architectures
	and languages using only XML and HTTP as a basis.

\item[WSDL] Web Services Definition Language, written in XML notation, is a language
	to describe types and message types a service provides and data exchanged in SOAP.
	WSDL's purpose is similar to IDL and it can be used to generate endpoint interfaces in different
	programming languages.

\end{description}



\chapter{System Overview}
\index{System Overview}

$Revision: 1.2 $

In this chapter, we examine the system architecture of the implementation
of the DMARF application and software interface design issues.

\section{Architectural Strategies}
\index{System Overview!Architectural Strategies}

The main principles are:

\begin{description}

\item [Platform-Independence] where one targets systems that are capable of running a JVM.

\item [Database-Independent API] will allow to swap database/storage engines on-the-fly. The appropriate
	adapters will be designed to feed upon required/available data source (binary, CSV file, XML, or SQL) databases.

\item [Communication Technology Independence] where the system design evolves such that any communication
	technologies adapters or plug-ins (e.g. {\rmi}, {\corba}, DCOM+, Jini, JMS, Web Services) can be added with little
	or no change to the main logic and code base.

\item [Reasonable Efficiency] where one architects and implements an efficient system, but will avoid advanced programming
	tricks that improve the efficiency at the cost of maintainability and readability.

\item [Simplicity and Maintainability] where one targets a simplistic and easy to maintain organization of the source.

\item [Architectural Consistency] where one consistently implements the chosen architectural approach.

\item [Separation of Concern] where one isolates separate concerns between modules and within modules to encourage re-use and code
simplicity.
\end{description}

\section{System Architecture}
\index{System Overview!System Architecture}

\subsection{Module View}
\index{System Overview!Module View}

\subsubsection{Layering}
\index{System Overview!Layering}

The DMARF application is divided into layers.
The top level has a front-end and a back-end.
The front-end itself exists on the client side
and on the server side. The client side is either
text-interactive, non-interactive 
client classes that connect and query the servers.
The front-end on the server side are the MARF pipeline
itself, the application-specific frontend, and pipeline stage services.
All pipeline stages somehow involved to the database and other storage
management subfunctions.
At the same time the services are a back-end for the client connecting in.

\subsection{Execution View}
\index{System Overview!Execution View}

\subsubsection{Runtime Entities}
\index{System Overview!Runtime Entities}

In the case of the DMARF application, there is hosting run-time environment
of the JVM and on the server side there must be the naming and implementation
repository service running, in the form of \tool{orbd} and \tool{rmiregistry}.
For the WS aspect of the application, there ought to be DNS running and
a web servlet container. The DBS uses Tomcat \cite{tomcat} as a servlet
container for MARF WS.
The client side for RMI and CORBA clients just requires a JRE (1.4 is the minimum).
The WS client in addition to JRE may require a servlet container environment
(here Tomcat) and a browser to view and submit a web form.
Both RMI and CORBA client and server applications are
stand-alone and non-interactive. A GUI is projected for the client
(and possibly server to administer it) in one of the follow up versions.

\subsubsection{Communication Paths}
\index{System Overview!Communication Paths}

It was resolved that the modules would all communicate through message passing
between methods. {\corba} is one of the networking technologies used for remote invocation.
{\rmi} is the base-line technology used for remote method calls. Further, a JAX-RPC over SOAP is used
for Web Services (while a more modern JAX-WS alternative to JAX-RPC was released, this project
still relies on JAX-RPC 1.1 as it's not using J2EE and the author found it simpler and faster
to use given the timeframe and more accurate tutorial and book material available).
All: {\rmi}, {\corba}, and WS influenced some technology-specific design decisions, but it
was possible to abstract them as RMI and CORBA ``agents'' and delegate
the business logic
to delegate classes 
enabling all three types of services to communicate in the future and implement transactions similarly.
Communication to the database depends on the storage manager (each terminal business logic
module in the classical MARF is a StorageManager).
Additionally, Java's reflection \cite{java-reflection} is used to discover instantiation communication
paths at run-time for pluggable modules.

\subsubsection{Execution Configuration}
\index{System Overview!Execution Configuration}

The execution configuration of the DMARF has to do with where its
\file{data/} and \file{policies/} directories are. The \file{data/} directory is always
local to where the application was ran from. In the case of WS, it has to be where Tomcat's
current directory is; often is in the \file{logs/} directory of \verb+${catalina.base}+.
The data directory
contains the service-assigned databases in the \file{XXX.gzbin} (generated on the first run
of the servers).
The ``XXX'' corresponds to the either training set or a module name that saved their state.
Next, \tool{orbd} keeps its data structures and logs in the \file{orb.db/} directory
also found in the current directory
Additionally, the RMI configuration for application's (both client and server)
policy files is located in \file{allallow.policy} (testing with all
permissions enabled).
As for the
WS, for deployment two directories \file{META-INF/} and \file{WEB-INF/}
are used. The former contains the Tomcat's contex file for deployment
that ought to be placed in \verb+${catalina.base}/conf/Catalina/localhost/+
and the latter typically goes to \file{local/marf} as the context
describes. It contains \file{web.xml} and other XML files prduced to
describe servlet to SOAP mapping when generating \file{.war} files
with \tool{wscompile} and \tool{wsdeploy}.

The build-and-run files include the Ant \cite{ant} \file{dmarf-build.xml}
and the GNU make \cite{gmake} \file{Makefile} files. The \file{Makefile}
is the one capable of starting the \tool{orbd}, \tool{rmiregistry}, the servers, and
the clients in various modes. The execution configuration targets primarily Linux FC4
platform (if one intends to use \tool{gmake}), but is not restricted to it.

A hosts configuration file \file{dmarf-hosts.properties} is used to tell
the services of how to initialize and where to find other services initially.
If the file is not present, the default of host for all is assumed to be
\url{localhost}.

\section{Coding Standards and Project Management}
\index{System Overview!Coding Standards and Project Management}
\label{sect:standards}

In order to produce higher-quality code, it was decided to normalize on Hungarian Notation coding
style used in {\marf} \cite{marfcoding}.
Additionally, \tool{javadoc} is used as a source code documentation style for
its completeness and the automated tool support.
CVS (\tool{cvs}) \cite{cvs} was employed in order to manage the source code, makefile,
and documentation revisions.

\section{Proof-of-Concept Prototype Assumptions}
\index{System Overview!Prototype Assumptions}

Since this is a prototype application within a timeframe of a course, some
simplifying assumptions took place
that were not a part of, explicit or implied,
of the specification.

\begin{enumerate}
\item
	There is no garbage collection done on the server side in
	terms of fully limiting the WAL size.

\item
	WAL functinality has not been at all implemented for
	the modules other than Classification.

\item
	MARF services does not implement nested transaction
	while pipelining.

\item
	Services don't intercommunicate (TCP or UDP) other than through
	the pipeline mode of operation.

\item
	No primary-backup or otherwise replication is present.

\end{enumerate}

\section{Software Interface Design}
\index{System Overview!Software Interface Design}

Software interface design comprises both user interfaces
and communication interfaces (central topic of this work) between modules.

\subsection{User Interface}
\index{System Overview!User Interface}

For the RMI and CORBA clients and servers there is a
GUI designed for status and control as time did not permit to properly integrate one. Therefore,
they use a command-line interface that is typically invoked from
a provided \file{Makefile}. GUI integration is projected in the near future.
See the interface prototypes in \xf{fig:speak-gui-client} and in \xf{fig:server-status-gui}.

\begin{figure}[hb]
	\includegraphics[width=\textheight,angle=90]{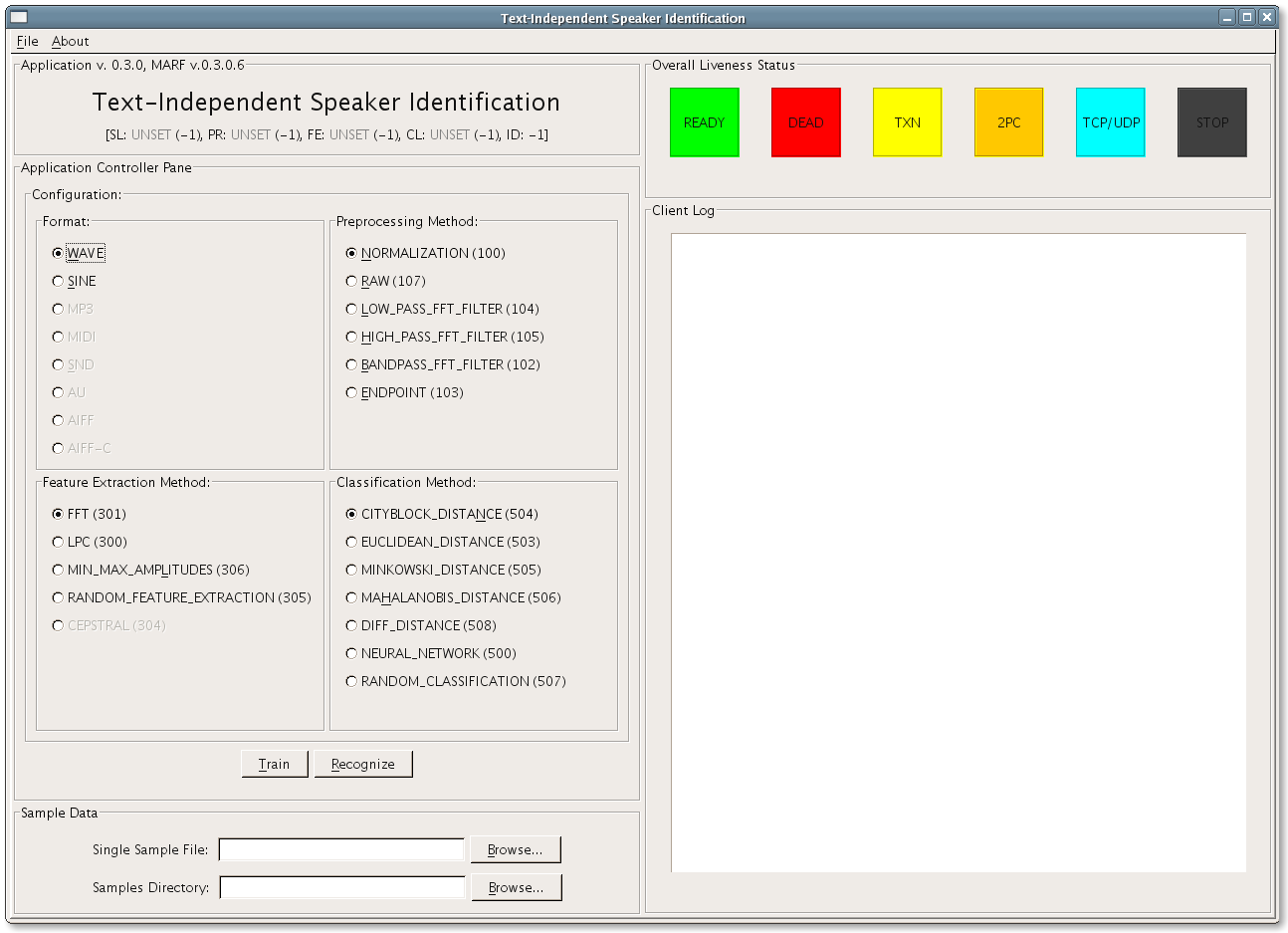}
	\caption{SpeakerIdenApp Client GUI Prototype}
	\label{fig:speak-gui-client}
\end{figure}

\begin{figure}[hb]
	\includegraphics[width=\textwidth]{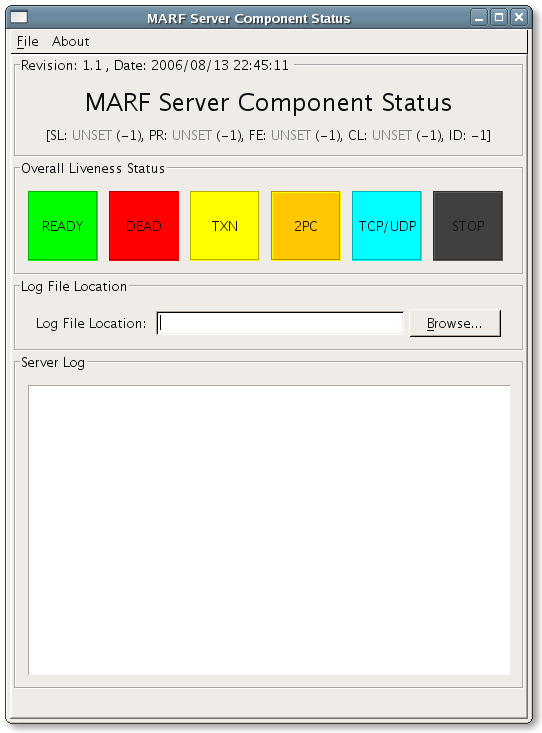}
	\caption{MARF Service Status Monitor GUI Prototype}
	\label{fig:server-status-gui}
\end{figure}

\subsection{Software Interface}
\index{System Overview!Software Interface}

Primary communication-related software interfaces are briefly
described below. A few other interfaces are omitted for brevity
(of storage and classical {\marf}).

\subsubsection{RMI}

The main RMI interfaces the RMI servants implement are
\api{ISpeakerIdentRMI}, \api{IMARFServerRMI}, \api{ISampleLoaderRMI}, \api{IPreprocessingRMI},
\api{IFeatureExtractionRMI}, and \api{IClassificationRMI}. They are located in the \apipackage{marf.server.rmi.*}
and \apipackage{marf.client.rmi.*}
packages. There also are the generated files off this interface for stubs
with \tool{rmic} and the servant implementation.

\subsubsection{CORBA}

The main CORBA IDL interfaces the servants implement are
\api{ISpeakerIdentCORBA}, \api{IMARFServerCORBA}, \api{ISampleLoaderCORBA}, \api{IPreprocessingCORBA},
\api{IFeatureExtractionCORBA}, and \api{IClassificationCORBA}. The IDL files are located in the \apipackage{marf.server.corba.*}
package and are called \file{MARF.idl} and \file{Frontends.idl}. There also are the generated files off this interface definition
for stub, skeleton, data types holders and helpers with \tool{idlj} and the servant implementation
and a data type adapter (described later).

\subsubsection{WS}

The main WS interface the WS ``servants'' (servlets) implement is
\api{ISpeakerIdentWS}, \api{IMARFServerWS}, \api{ISampleLoaderWS}, \api{IPreprocessingWS},
\api{IFeatureExtractionWS}, and \api{IClassificationWS}.
They are located in the \apipackage{marf.server.ws.*}.
There are also the generated files off this interface for stub and skeleton
serializers and builders for each method and non-primitive data type of Java
with \tool{wscompile} and \tool{wsdeploy} and the ``servant'' implementations.
There are about 8 files generated for SOAP XML messages per method or a data type
for requests, responses, faults, building, and serialization.

\subsubsection{Delegate}

The DMARF is flexible here and allows any delegate implementation as long
as \api{IDelegate} in \apipackage{marf.net.server.delelegates} is
implemented. A common implementation of it is also there provided with
the added value benefit that all three types of servants of the above can use
the same delegate implementations and therefore can share all of
functionality, transactions, and communication.

\subsection{Hardware Interface}
\index{System Overview!Hardware Interface}

Hardware interface is fully abstracted by the JVM and
the underlying operating system making the DMARF application
fully architecture-independent. The references to STDOUT
and STDERR (by default the screen or a file) are handled
through the \api{System.out} and \api{System.err} streams.
Likewise, STDIN (by default associated with keyboard) is
abstracted by Java's \api{System.in}.


\chapter{Detailed System Design}
\index{Detailed System Design}
\index{Design}

$Revision: 1.2 $

This chapter briefly presents the design considerations
and assumptions in the form of directory structure, class diagrams as well as
storage organization.

\section{Directory and Package Organization}
\index{Directory Structure}
\index{Package Organization}

In this section, the directory structure is introduced.
Please note that {\java}, by default, converts sub-packages
into subdirectories, which is what we see in \xf{fig:packages}.
Please also refer to \xt{tab:folders:explanation} and \xt{tab:marf-packages} for description of
the data contained in the directories and the package organization, respectively.

\begin{figure}[hb]
	\includegraphics[totalheight=\textheight]{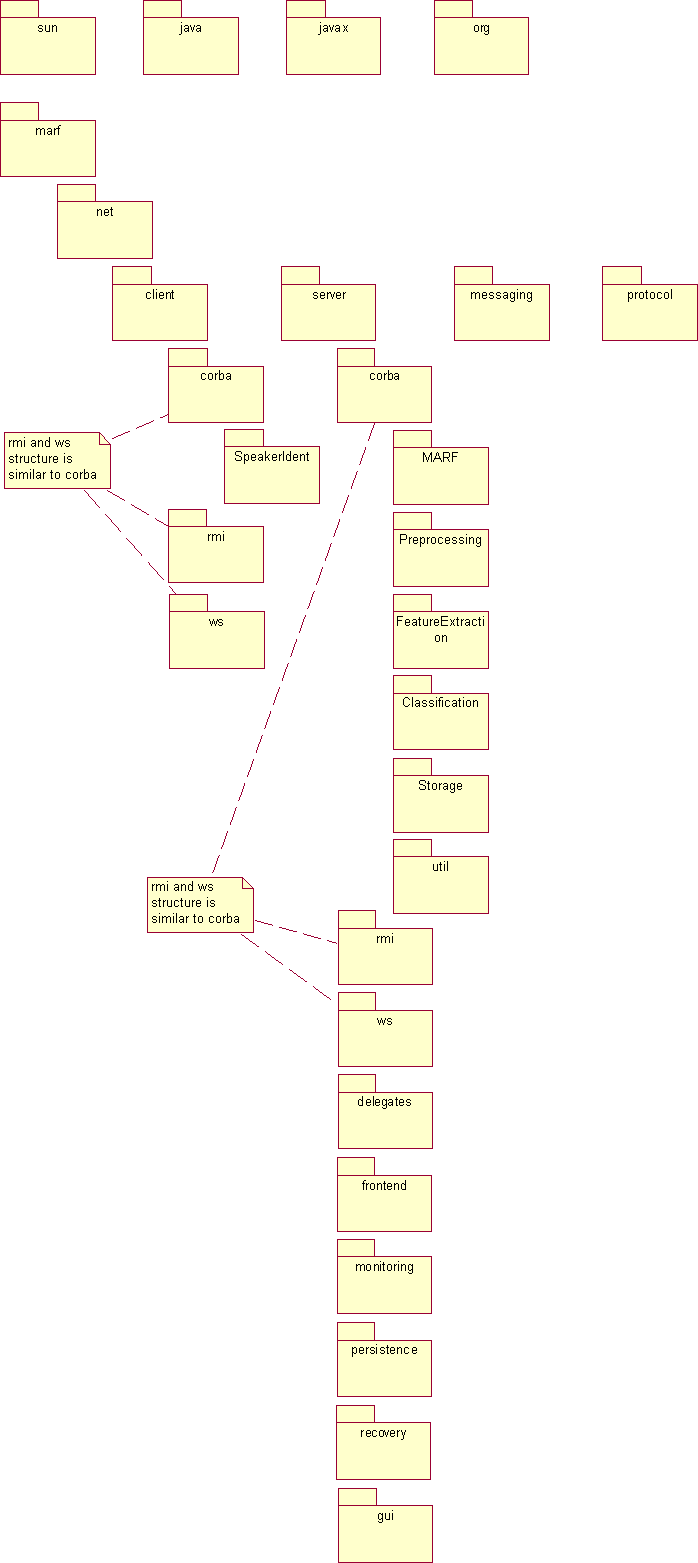}
	\caption{Package Structure of the Project}
	\label{fig:packages}
\end{figure}

\begin{table}
\begin{tabular}{|l|p{3.5in}|}
	\hline
	\textbf{Directory}	& \textbf{Description} \\ \hline
	\file{bin/}			& compiled class files are kept here. The sub-directory structure
						  mimics the one of the \file{src/}. \\ \hline
	\file{data/}		& contains the database as well as \file{stocks} file.\\ \hline
	\file{logs/}		& contains the client and server log files as ``screenshots''.\\ \hline
	\file{orb.db/}		& contains the naming database as well as logs for the \tool{orbd}.\\ \hline
	\file{doc/}			& project's API and manual documentation (well and theory as well). \\ \hline
	\file{lib/}			& meant for libraries, but for now there is none\\ \hline
	\file{src/}			& contains the source code files and follows the described package hierarchy.\\ \hline
	\file{dist/}		& contains distro services \file{.jar} and \file{.war} files.\\ \hline
	\file{policies/}	& access policies for the RMI client and server granting various permissions.\\ \hline
	\file{META-INF/}	& Tomcat's context file (and later manifest) for deployment \file{.war}.\\ \hline
	\file{WEB-INF/}		& WS WSDL servlet-related deployment information and classes.\\ \hline
\end{tabular}
\caption{Details on Main Directory Structure}
\label{tab:folders:explanation}
\end{table}

\begin{table}
\begin{tabular}{|l|p{3.5in}|}
	\hline
	\textbf{Package}			& \textbf{Description}  \\ \hline
	\apipackage{marf}		& root directory of the {\marf} project; below are the packages mostly pertinent to the DMARF\\ \hline
	\apipackage{marf.net.*.*}	& {\marf}'s directory for the some generic networking stuff\\ \hline
 	\apipackage{marf.net.client}	& client application code and subpackages\\ \hline
	\apipackage{marf.net.client.corba.*}	& Distributed MARF CORBA clients\\ \hline
	\apipackage{marf.net.client.rmi.*}	& Distributed MARF RMI clients\\ \hline
	\apipackage{marf.net.client.ws.*}	& Distributed MARF WS clients\\ \hline
	\apipackage{marf.net.messenging.*}	& Reserved for message-passing protocols\\ \hline
	\apipackage{marf.net.protocol.*}	& Reserved for other protocols, like two-phase commit\\ \hline
 	\apipackage{marf.net.server.*}	& main server code and interfaces is placed here\\ \hline
 	\apipackage{marf.net.server.rmi.*}	& RMI-specific sevices implementation\\ \hline
 	\apipackage{marf.net.server.corba.*}	& CORBA-specific sevices implementation\\ \hline
 	\apipackage{marf.net.server.ws.*}	& WS-specific sevices implementation\\ \hline
 	\apipackage{marf.net.server.delegates.*}	& service delegate implementations are here\\ \hline
 	\apipackage{marf.net.server.frontend.*}	& root of the service front-ends\\ \hline
 	\apipackage{marf.net.server.frontend.rmi.*}	& RMI-specific service front-ends\\ \hline
 	\apipackage{marf.net.server.frontend.corba.*}	& CORBA-specific service front-ends\\ \hline
 	\apipackage{marf.net.server.frontend.ws.*}	& WS-specific service front-ends\\ \hline
 	\apipackage{marf.net.server.frontend.delegates.*}	& service front-ends delegate implementations\\ \hline
 	\apipackage{marf.net.server.gossip}	& reserved for the gossip replication implementation\\ \hline
 	\apipackage{marf.net.server.gui}	& server status GUI\\ \hline
 	\apipackage{marf.net.server.monitoring}	& reserved for various service monitors and their bootstrap\\ \hline
 	\apipackage{marf.net.server.persistence}	& reserved for WAL and Transaction storage management\\ \hline
 	\apipackage{marf.net.server.recovery}	& reserved for WAL recovery and logging\\ \hline
	\apipackage{marf.Storage}	& {\marf}'s storage-related utility classes\\ \hline
	\apipackage{marf.util}	& {\marf}'s general utility classes (threads, loggers, array processing, etc.)\\\hline
	\apipackage{marf.gui}	& general-purpose GUI utilities that to be used in the MARF apps, clients, and server status monitors\\ \hline
\end{tabular}
\caption{D{\marf}'s Package Organization}
\label{tab:marf-packages}
\end{table}

\clearpage
\section{Class Diagrams}
\index{Design!Class Diagrams}

At this stage, the entire design is summarized in five class diagrams representing the major modules
and their relationships. The diagrams of the overall architecture and its storage subsystem
are in \xf{fig:general} and \xf{fig:storage} respectively. Then, some details on {\corba}, {\rmi}, and WS
implementations are in \xf{fig:corba}, \xf{fig:rmi}, and \xf{fig:ws} respectively.
Please locate the detailed description
of the modules in the generated API HTML off javadocs or the javadoc
comments themselves in the \file{doc/api} directory. Some of the
description appears here as well in the form of interaction between
classes.

\begin{figure}
\includegraphics[width=1.1\textwidth,totalheight=\textheight]{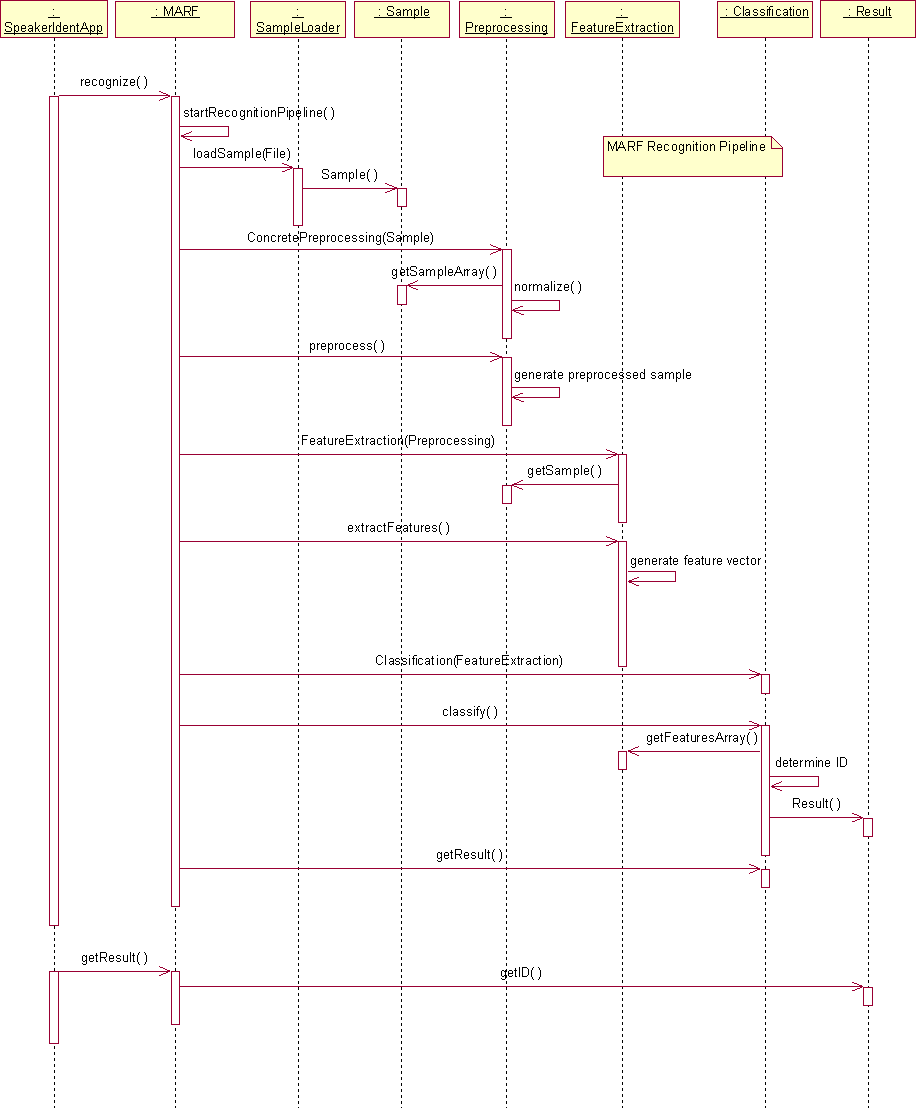}
\caption{Sequence Diagram of the Pipeline Of Invocations}
\label{fig:pipeline-seq}
\end{figure}

\begin{figure}
\includegraphics[width=1.1\textwidth,totalheight=\textheight]{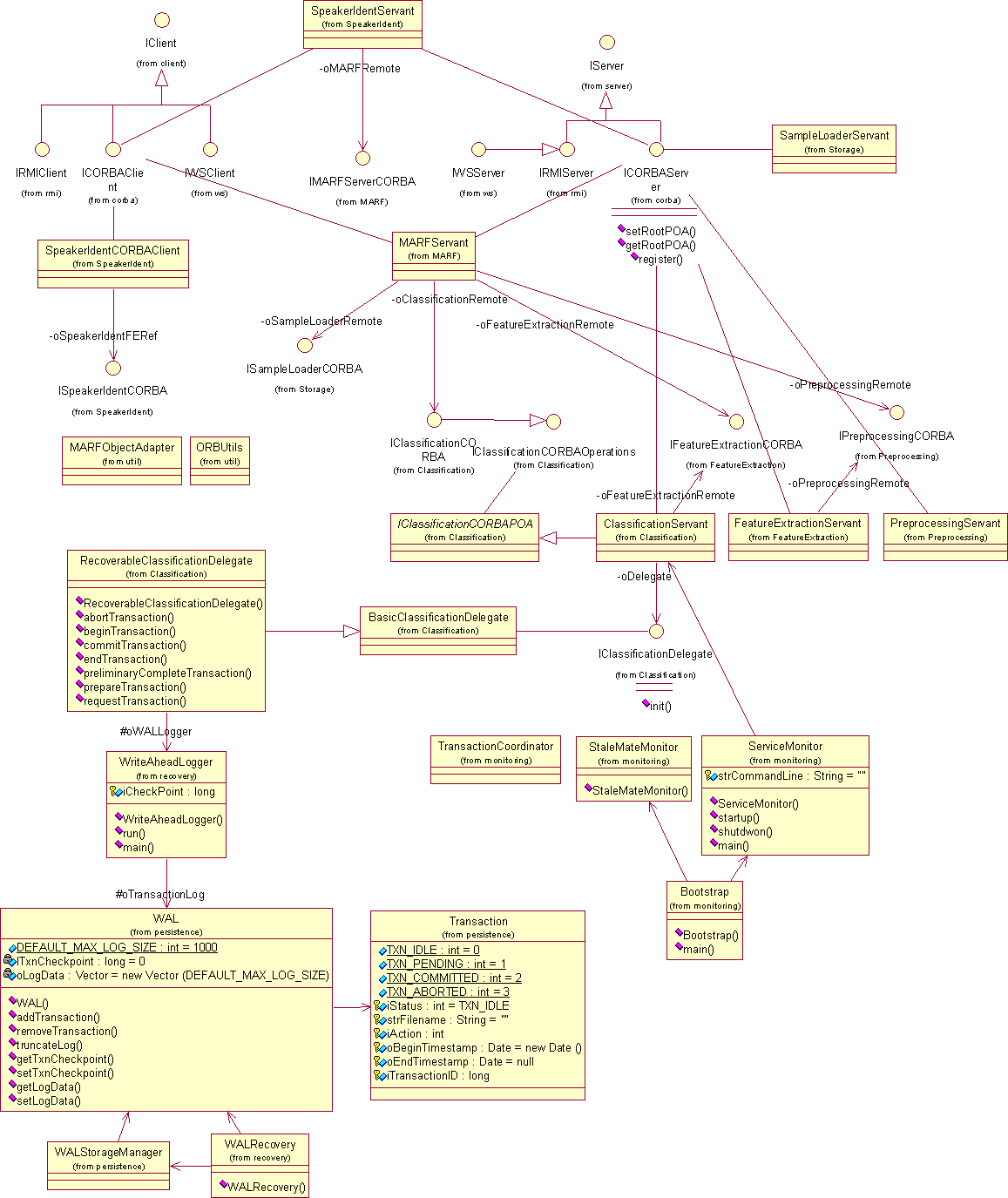}
\caption{General Architecture Class Diagram of marf.net}
\label{fig:general}
\end{figure}

At the beginning of the hierarchy are the \api{IClient} and \api{IServer} are independent of a communication technology
type of interfaces that ``mark'' the would-be classes of either type. This is
design of a system where one will be able to pick and choose either
manually or automatically which communication technologies to use. These
interfaces are defined in the \apipackage{marf.net} and used in reflection instantiation
utils.


Next, the hierarchy branches to the {\corba}, {\rmi}, and WS marked-up sever and client interfaces,
\api{ICORBAServer}, \api{ICORBAClient}, \api{IRMIServer} and \api{IRMIClient}, \api{IWSServer} and \api{IWSClient}.
The specificity of the
\api{IRMIServer} that it extends the \api{Remote} interface required
by the {\rmi} specification. The \api{ICORBAServer} allows to set and get the
root POA. And the \api{IWSServer} allows setting and getting an in-house made
\api{RemoteObjectReference} (which isn't true object reference as in RMI or
CORBA, but incapsulates the necessary service location information).

Then, the diagram shows only the CORBA details (and RMI and WS are similar, but
the diagram is already cluttered, so they were omitted). Then the diagram
shows all six servants and their relationships with the interfaces
as well as blending in WAL logging and transaction recovery. There some
monitoring modules designed as well.

The clients for the respective technologies are in the
\apipackage{marf.net.client.corba},
\apipackage{marf.net.client.rmi}, and
\apipackage{marf.net.client.ws}
packages.


When implementing the CORBA services, a data type adapter had to be made
to adapt certain data structures that came from \file{MARF.idl} to the common
storage data structures (e.g. \api{Sample}, \api{Result}, \api{CommunicationException}, \api{ResultSet}, etc.).
Thus, the \api{MARFObjectAdapter} class was provided to adapt these data
structured back and forth with the generic delegate when needed.

The servers for the respective technologies are in the
\apipackage{marf.net.server.corba},
\apipackage{marf.net.server.rmi}, and
\apipackage{marf.net.server.ws}
packages.

Finally, on the server side, the \api{RecoverableClassificationDelegate} interacts with the \api{WriteAheadLogger}
for transaction information. The storage manager here serializes
the WAL entries

\begin{figure}
	\includegraphics[width=\textheight, totalheight=\textwidth, keepaspectratio=true, angle=90]{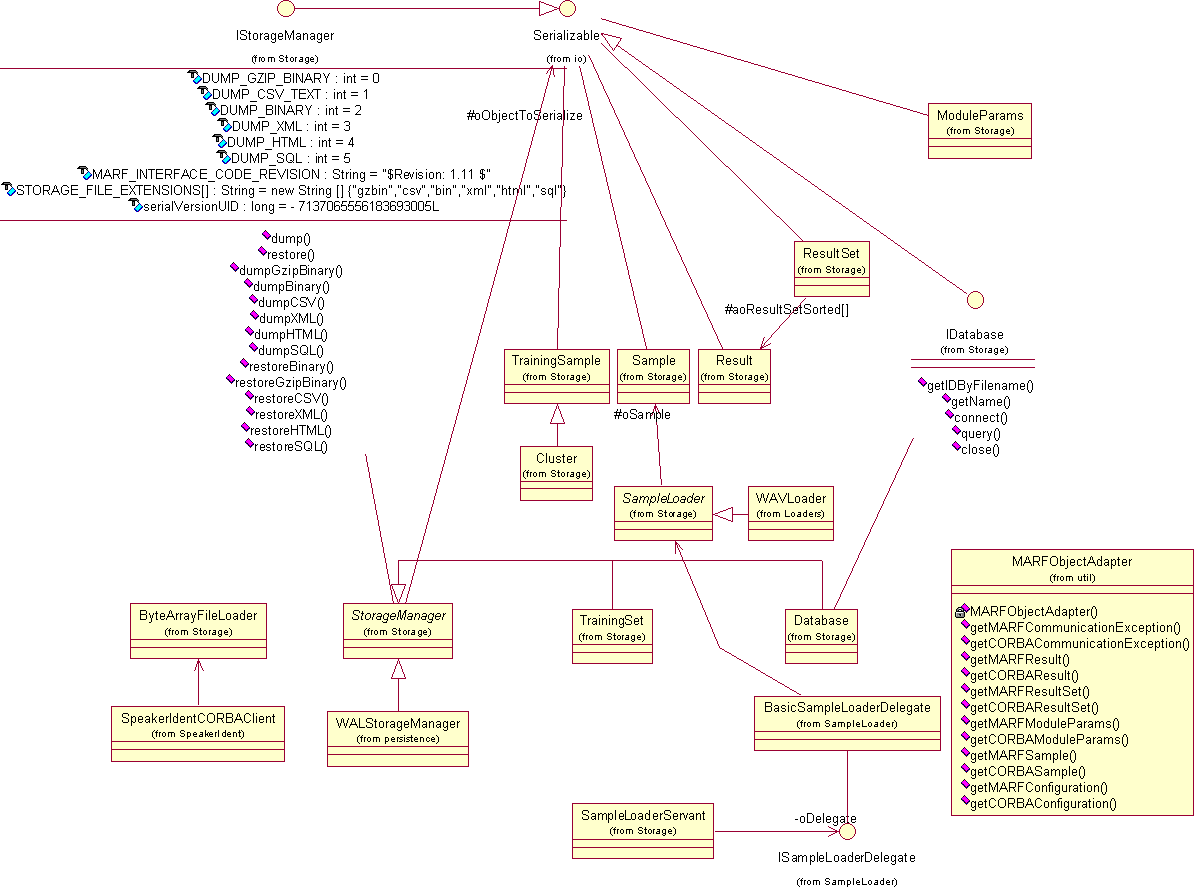}
	\caption{Storage Class Diagram}
	\label{fig:storage}
\end{figure}

More design details are revealed in the class diagram of the storage-related aspects in
\xf{fig:storage}. The \api{Database} contains stats of classificatio and is only
written by the SpeakerIdent front-end. All, \api{Database}, \api{Sample}, \api{Result},
and \api{ResultSet} and \api{TrainingSet} implement \api{Serializable}
to be able to be stored on disk or transferred over a network.

The serialization of the \api{WAL} instance into the file
is handled by the \api{WALStorageManager} class.
The \api{IStorageManager} interface
and its most generic implementation \api{StorageManager} also come from my {\marf}'s
\apipackage{marf.Storage} package. The \api{StorageManager} class provides
the implementation of serialization of classes in plain binary as well as
compressed binary formats. (It also has facilities to plug-in other storage
or output formats, such as CSV, XML, HTML, and SQL, which derivatives
must implement if they wish.

\section{Data Storage Format}
\index{Design!Data Storage Format}
\index{Data Storage Format}

This section is about data storage issues and the details
on the chosen underlying implementation and ways of addressing
those issues.
For the details on the classical MARF storage susbsystem
please refer to the Storage chapter in \cite{marf}.

%

\subsection{Log File Format}
\index{Data Storage Format!Log File Format}

The log is saved in the \file{module-technology.log} files for the
server and client respectively
in the application's current directory.
As of this version, the
file is produced with the help of the \api{Logger} class that is in \apipackage{marf.util}.
(Another logging facility that was considered but not yet
only used in WS with Tomcat is the Log4J\index{Libraries!Log4J} tool \cite{log4j}, which has a full-fledged logging engine.)
The log file produced by \api{Logger} has a classical format of ``\verb+[ time stamp ]: message+''.
The logger intercepts all attempts to write to STDOUT or STDERR and makes
a copy of them to the file. The output to SDTOUT and STDERR is also preserved.
If the file remains between different runs, the log data is appended.

\section{Synchronization}
\index{Synchronization}

The notion of synchronization is crucial in an application that allows
access to a shared resource or a data structure by multiple clients.
This includes our DMARF. At the server side the synchronization must be
maintained when the \api{Database} or \api{TrainingSet} objects are accessed through the
server possibly by multiple clients. The way it is
implemented in this version, the \api{Database} class becomes its own
object monitor and all its relevant methods are made \api{synchronized},
thus locking entire object while it's accessed by a thread thereby
providing data integrity. The whole-instance locking maybe a bit
inefficient, but can be careful re-done by only marking some
critical paths only and not the entire object.

Furthermore, multiple server keep a copy of their own dats structures,
including stock data, making it more concurrent. On top of that,
the WS, RMI, and CORBA brokers act through a delegate implementation
allowing to keep all the synchronization and business logic in one
place and decouple communication from the logic.
The rest is taken care of by the WAL.

\section{Write-Ahead Logging and Recovery}

The recovery log design is based on the principle of the write-ahead
logging. This means the transaction data is written first to the log,
and upon successful return from writing the log, the transaction is
committed.

Checkpointing is done periodically of flushing all the transactions
to the disk with the record of the latest committed transaction ID
as a checkpoint. In the even of crash, upon restart, the WAL is read
and the object states are recovered from the latest checkpoint.

The design of the WAL algorithm in DMARF is modified such that
the logged transaction data contains the ``before'' and ``after''
snapshots of the object in question (a training set, message, or
the whole module itself). In part this is due to the fact that
the transactions are wrapped around classical business logic, that does
alter the objects on disk, so in the even of a failure the ``before''
snapshot is used to revert the object state on disk the way it was
back before the transaction in question began.

WAL grows up to a certain number of committed transactions.
Periodic garbage collection on WAL and checkpointing are performed.
At the garbage collection oldest aborted transactions are removed
as well as up to a 1000 committed transactions. WAL can be
periodically backed up, shipped to another server for replication,
or point-in-time recovery (PITR) and there are timestamps associated
with each serialized transaction.

In most part, WAL is pertinent to the Classification service as this
is where most of writes are done during the training phase (in the
classification phase it is only reading). Sample loading, preprocessing,
and feature extraction services can also perform intermediate writes
if asked, but most of the time they crunch the data and pass it around.
The classification statistics is maintained at the application-specific
front-end for now, and there writes are serialized.

\section{Replication}

The replication is done by either the means of WAL (ship over WAL
to another host and ``replay'' it along certain timeline). Another
way is lazy update though the gossip architecture among replica.
Delegates broadcast ``whoHas(config)'' requests before computing anything
theselves; if shortly after no response received, the delegate issuing
the request starts to compute the configuration itself, else a transfer
is initiated from another delegate that have computed an identical configuration.


\chapter{Testing}
\index{Testing}

The conducted testing of (mostly CORBA) pipeline including
single training test and a batch training on maximum four
computers in separate buildings. \file{Makefile} and
\file{batch.sh} serve this purpose. If you intend to
use them, make sure you have the server jars in \file{dist/}
and properly configured \file{dmarf-hosts.properties}.

The tests were quite successful and terminating any of the
service replicas and restarting it resumed normal operation
of the pipeline in the batch mode. There more thorough testing
is to be conducted as the project evolves from a proof-of-concept
to a cleaner solution.


\chapter{Conclusion}

$Revision: 1.2 $

Out of the three main distributed technologies learnt and used through the course
({\rmi}, {\corba}, and Web Services) to implement
the MARF services, I managed to implement all three.

The Java {\rmi} technology seems to be the lowest-level of remote method
invocation tools for programmers to use. Things like Jini, JMS tend to be
more programmer-friendly. Additional limitation that {\rmi} has as
the requirement of the remote methods to throw the \api{RemoteException}
and when generating stubs RMI-independent interface hierarchy does work.

A similar problem exists for {\corba}, which generates even CORBA-specific
data structures from the struct definitions that cannot be easily linked
to the data structures used elsewhere throughout the program through
inheritance or interfaces.

The WS implementation from the Java-endpoint provided interface and
and a couple of XML files was a natural extension of {\rmi} implementation
{\em but} with somewhat different semantics. The implementation aspect
was not hard, but the deployment within a servlet container
and WSDL compilation were a large headache.

However, highly modular design allowed swapping module implementations from one
technology to another if need be making it very extensible by the means of
delegating the actual business logic to the a delegate classes. As an added
bonus of that implementation, RMI, CORBA, WS services can communicate
through TCP or UDP and do transaction.
Likewise, all the synchronization efforts are undertaken by the delegate
and the delegate is the single place to fix is there is something broken.
Aside from the delegate class, a data adapter class for CORBA also contributes
here to translate the data structures.

\section{Summary of Technologies Used}

The following were the most prominent technologies used throughout the implementation of the project:

\begin{itemize}
\item J2SE (primarily 1.4)
\item Java IDL \cite{java-corba-idl}
\item Java RMI \cite{java-rmi}
\item Java WS with JAX-RPC \cite{java-webservices}
\item Java Servlets \cite{servlets}
\item Java Networking \cite{java-networking}
\item Eclipse IDE \cite{eclipse}
\item Apache Ant \cite{ant}
\item Apache Jakarta Tomcat 5.5.12 \cite{tomcat}
\item GNU Make \cite{gmake}
\end{itemize}



\section{Future Work and Work-In-Progress}

Extend the remote framework to include other communication technologies (Jini, JMS, DCOM+, .NET Remoting)
in communication-independent fashion and transplant
that all for use in {\marf} \cite{marf}. Additionally, complete
application GUI for the client and possibly server implementations.
Finally, complete the advanced features of distributed systems such
as disaster recovery, fault tolerace, high availability and replication,
and others with great deal of thorough testing.

\section{Acknowledgments}

\begin{itemize}
\item The authors of the Java RMI \cite{java-rmi}, Java IDL \cite{java-corba-idl}, Java Web Services \cite{java-webservices}
reference material from Sun.
\item The authors of the textbook \cite{disysconcepts05}.
\item Dr. Rajagopalan Jayakumar for the Distributed Systems Design Course
\item Dr. Peter Grogono for {\LaTeX} introductory tutorial \cite{grogono2001}
\item Nick Huang, the TA
\end{itemize}


\addcontentsline{toc}{chapter}{References}

\bibliography{sdd}
\bibliographystyle{alpha}







\printindex

\end{document}